\documentclass[final]{cvpr}

\usepackage{times}
\usepackage{epsfig}
\usepackage{graphicx}
\usepackage{subcaption}
\usepackage{multirow}
\usepackage{multicol}
\usepackage{pifont}
\newcommand{\cmark}{\ding{51}}%
\newcommand{\xmark}{\ding{55}}%
\usepackage{amsmath}
\usepackage{amssymb}
\usepackage{mathtools}
\usepackage[super]{nth}

\usepackage[pagebackref=true,breaklinks=true,colorlinks,bookmarks=false]{hyperref}

\def\cvprPaperID{****} 
\def\confYear{CVPR 2021}

\begin{document}

\title{Real-Time Anchor-Free Single-Stage 3D Detection with IoU-Awareness}
\author{
Runzhou Ge$^{*}$ \quad Zhuangzhuang Ding$^{*}$ \quad Yihan Hu$^{*}$  \\
Wenxin Shao \quad Li Huang \quad Kun Li \quad Qiang Liu \\
Horizon Robotics\\
{\tt\small \{runzhouge, dinghouzx, yihan.hu96\}@gmail.com} 
}

\maketitle

\newcommand\blfootnote[1]{%
  \begingroup
  \renewcommand\thefootnote{}\footnote{#1}%
  \addtocounter{footnote}{-1}%
  \endgroup
}
\blfootnote{
This is the report for the \nth{1} Place Solutions to the Real-time 3D Detection and the ``Most Efficient Model'' of the Waymo Open Dataset Challenges 2021.
}
\blfootnote{
$\prescript{*}{}{}$These authors contributed equally to this work.
}

\begin{abstract}
In this report, we introduce our winning solution to the Real-time 3D Detection and also the ``Most Efficient Model'' in the Waymo Open Dataset Challenges at CVPR 2021. Extended from our last year's award-winning model AFDet, we have made a handful of modifications to the base model, to improve the accuracy and at the same time to greatly reduce the latency. The modified model, named as AFDetV2, is featured with a lite 3D Feature Extractor, an improved RPN with extended receptive field and an added sub-head that produces an IoU-aware confidence score. These model enhancements, together with enriched data augmentation, stochastic weights averaging, and a GPU-based implementation of voxelization, lead to a winning accuracy of 73.12 mAPH/L2 for our AFDetV2 with a latency of 60.06 ms, and an accuracy of 72.57 mAPH/L2 for our AFDetV2-base, entitled as the ``Most Efficient Model'' by the challenge sponsor, with a winning latency of 55.86 ms.
\end{abstract}



\section{Introduction}

The Waymo Open Dataset Challenges at CVPR 2021 are the most exciting competitions in the field of autonomous driving. The Waymo Open Dataset~\cite{sun2020scalability} and the Waymo Open Motion Dataset~\cite{ettinger2021large} are datasets with high-quality data collected from both LiDAR and camera sensors in real self-driving scenarios, which have enabled many new exciting research. The Real-time 3D Detection challenge requires an algorithm to detect the 3D objects of interest as a set of 3D bounding boxes within 70 ms per frame. The model with the highest performance (in mAPH/L2) while satisfying this real-time requirement wins this challenge. The model with the lowest latency and mAPH/L2 $>$ 70 is given the title of ``Most Efficient Model''.

\begin{figure}[t]
\centering
\begin{subfigure}[b]{.45\columnwidth}
\centering
\includegraphics[width=.7\linewidth]{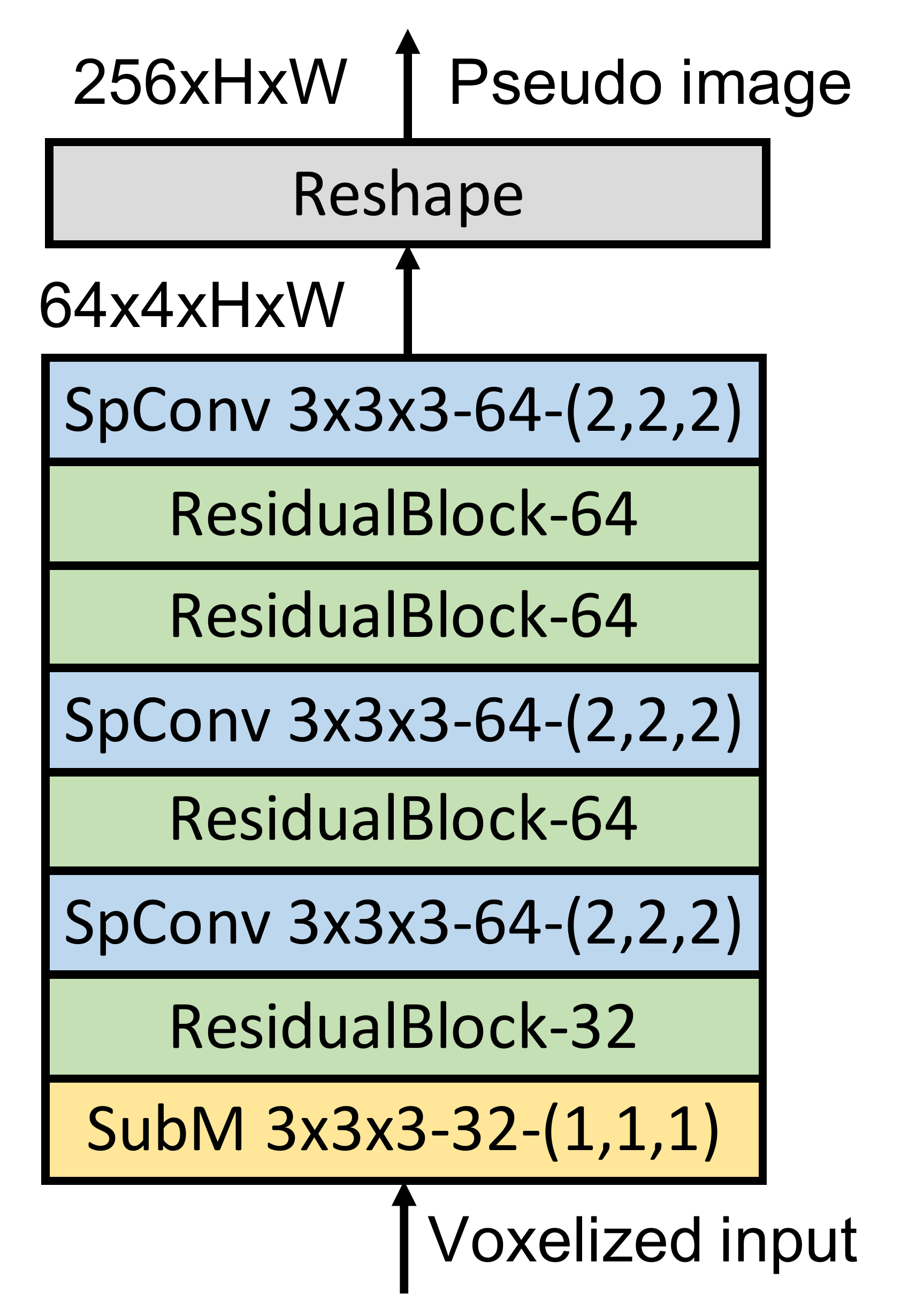}
\caption{3D Feature Extractor}\label{fig:feature_extractor}
\end{subfigure}
\hfill
\begin{subfigure}[b]{.45\columnwidth}
\centering
\includegraphics[width=.7\linewidth]{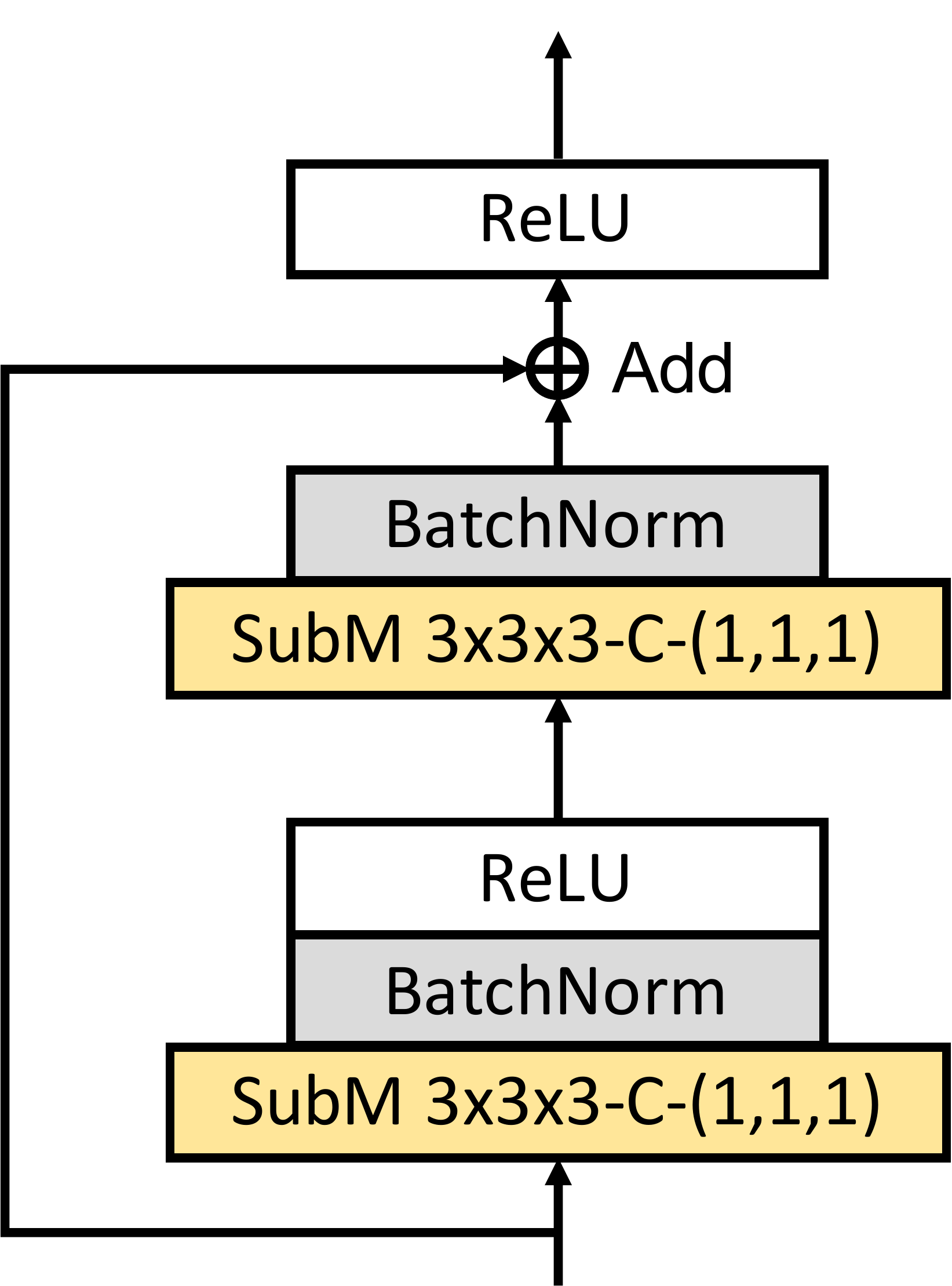}
\caption{Residual block}\label{fig:residual_block}
\end{subfigure}
\caption{(a) The structure of the lite 3D Feature Extractor. ``SubM'' stands for sub-manifold sparse convolutional layer~\cite{graham2017submanifold} and ``SpConv'' stands for 3D sparse convolutional layer~\cite{yan2018second}. The format of the layer setting follows ``kernel size-channels-(strides)'', \ie $k_W \times k_H \times k_D$-$C$-($s_H$,$s_W$,$s_D$). For residual blocks, only channels are shown. After the backbone, channel dimension and $D$ dimension are reshaped together to form a pseudo image. (b) The structure of a residual block with $C$ channels.}
\label{fig:backbone}
\end{figure}


\begin{figure*}
\begin{subfigure}[b]{.27\textwidth}
\centering
\includegraphics[width=\linewidth]{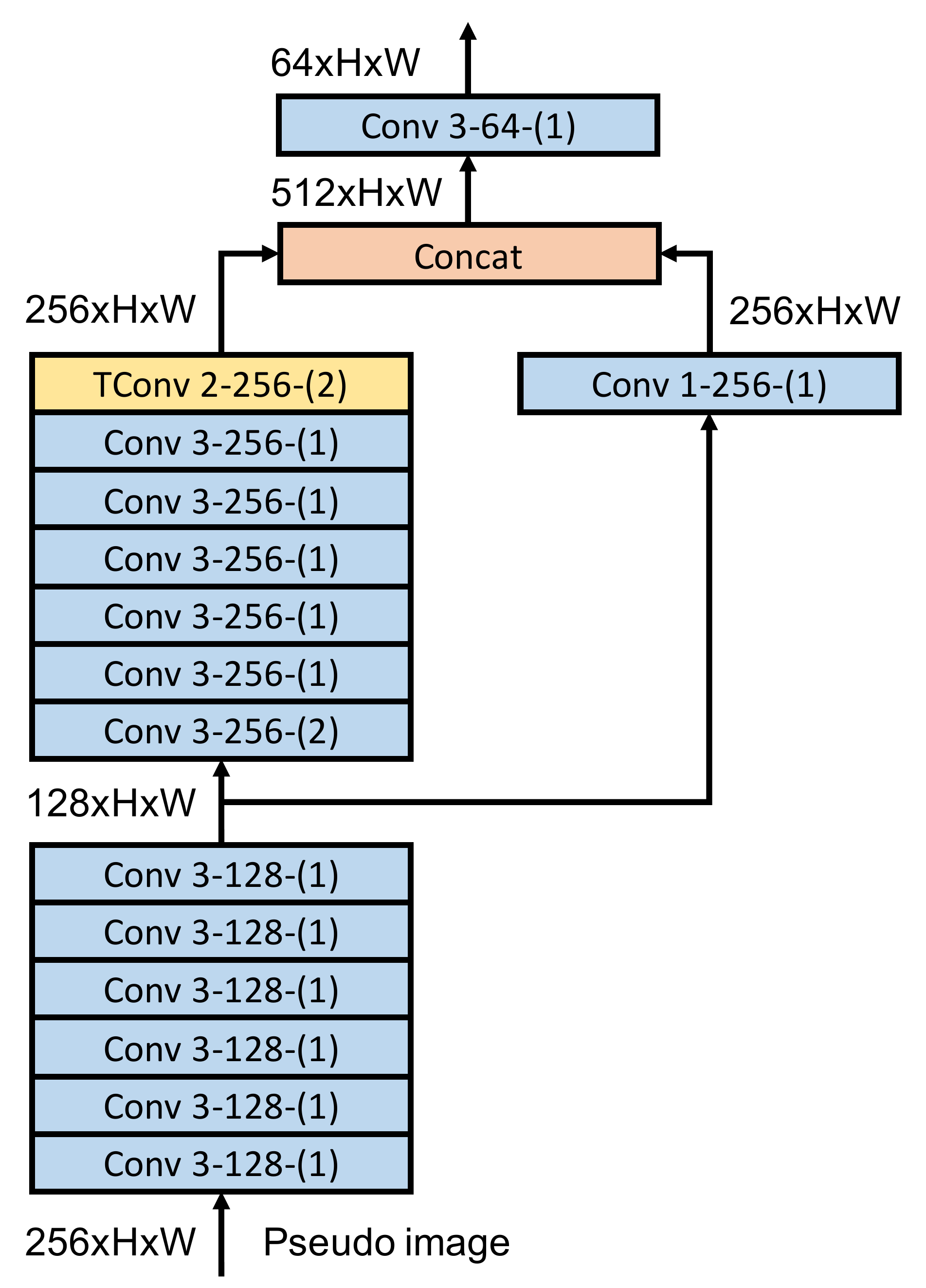}
\caption{RPN baseline}\label{fig:RPN1}
\end{subfigure}
\hfill
\begin{subfigure}[b]{.27\textwidth}
\centering
\includegraphics[width=\linewidth]{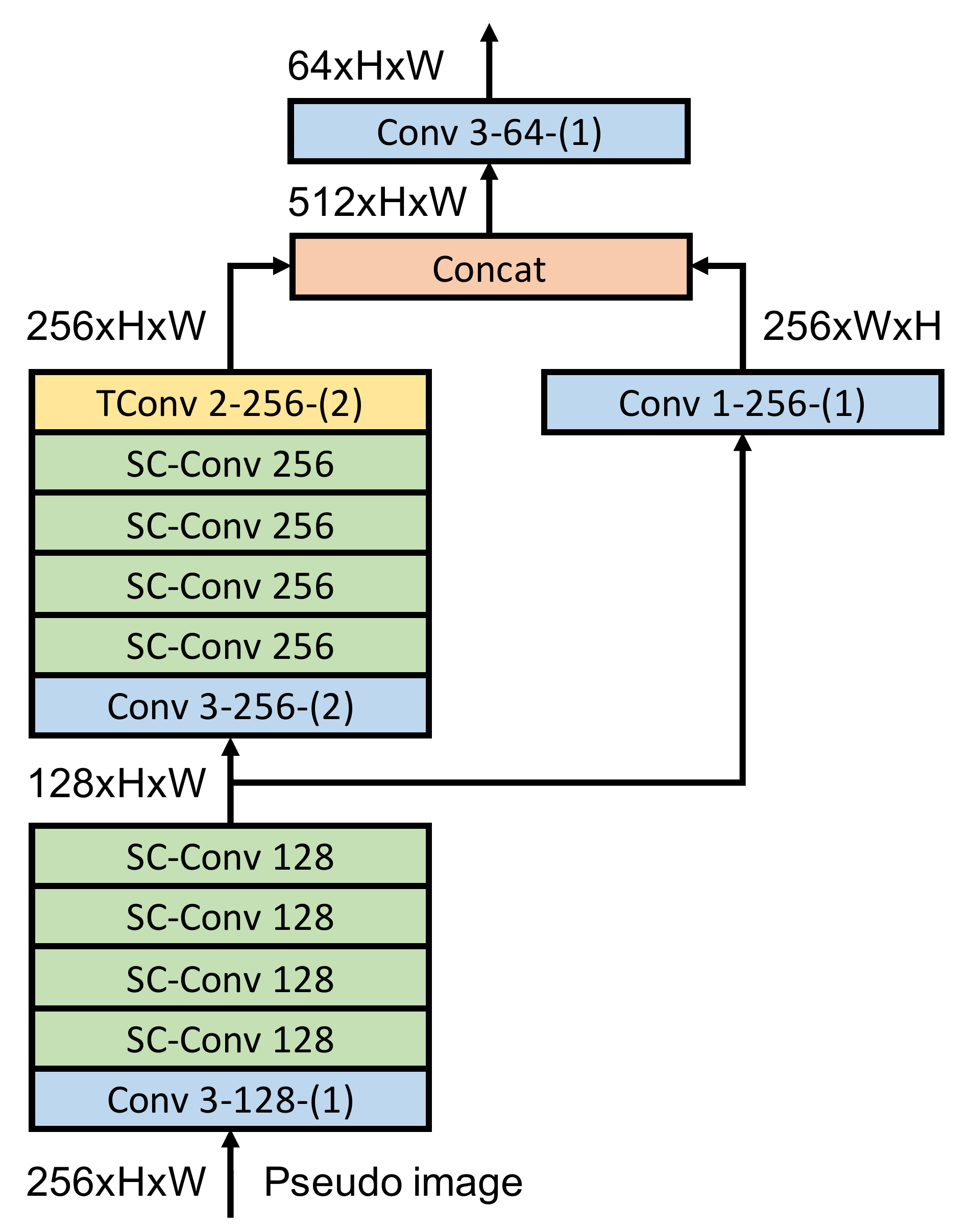}
\caption{SC-Conv RPN}\label{fig:RPN2}
\end{subfigure}
\hfill
\begin{subfigure}[b]{.36\textwidth}
\centering
\includegraphics[width=\linewidth]{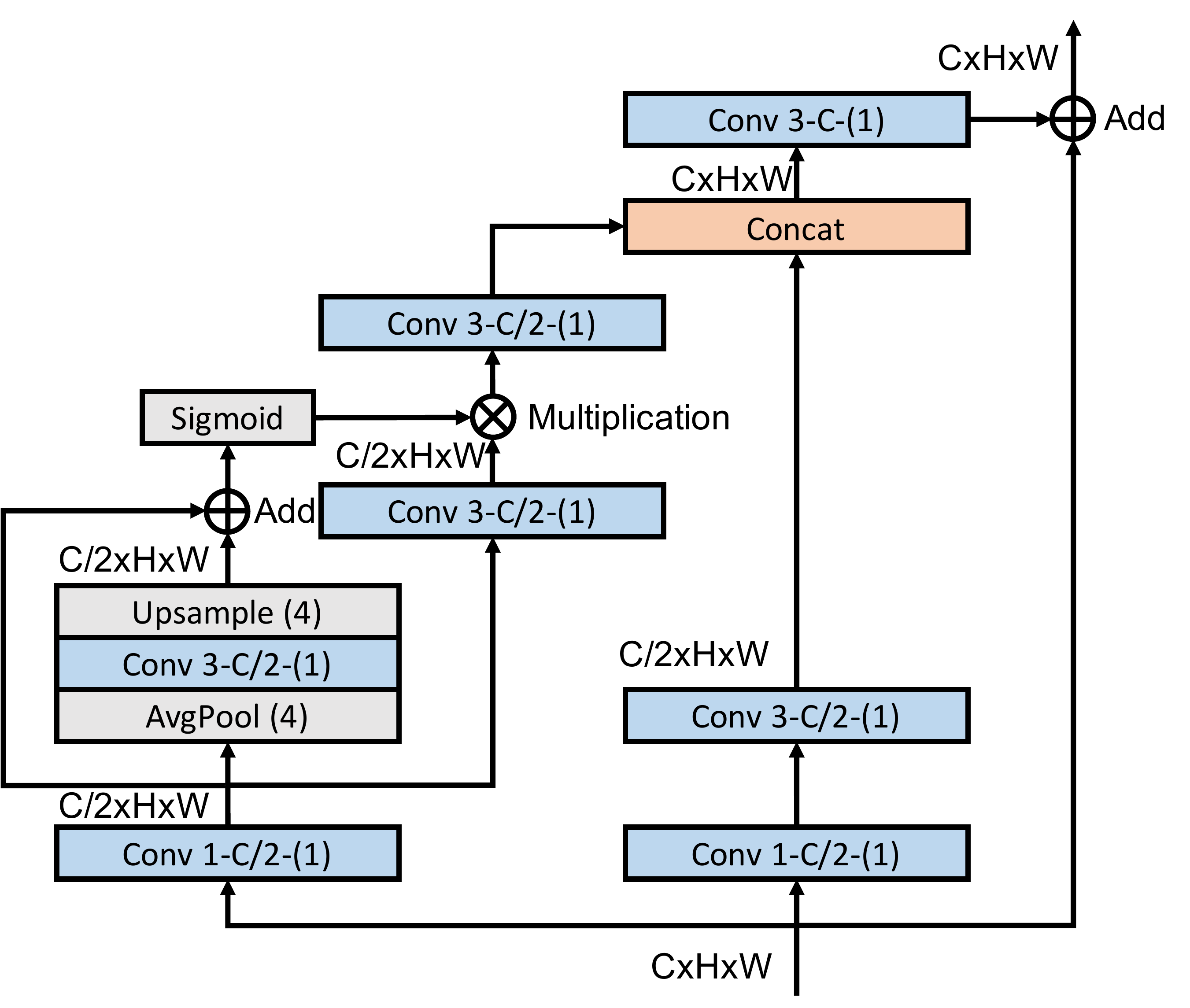}
\caption{SC-Conv module}\label{fig:SCConv}
\end{subfigure}
\caption{(a) and (b) denote the RPN baseline and the self-calibrated convolution (SC-Conv) RPN. ``Conv'' stands for convolutional layer and ``TConv'' stands for transposed convolutional layer. The format of the layer setting follows ``kernel size-channels-(strides)'', \ie $k$-$C$-($s$). ``SC-Conv'' stands for SC-Conv module and only channels are shown. (c) gives the detailed structure of the SC-Conv module.}
\label{fig:RPN}
\end{figure*}


\section{Methods}
In this section, we present the details of our network used in the challenge. The overall network structure follows our previous work AFDet~\cite{ge2020afdet}, which is a one-stage and anchor-free 3D point cloud detector. The whole network consists of four modules: Point Cloud Voxelization, 3D Feature Extractor, Region Proposal Networks (RPN) and Anchor-free Detection Head. We have made significant improvements to the model structure to deal with the new requirements of this challenge, especially the low latency requirement. We explored each module and its structure thoroughly in order to get the best performance while keeping the model structure simple.
\subsection{Point Cloud Voxelization}
Point cloud input is first voxelized~\cite{zhou2018voxelnet, zhu2019class}  into small voxels across $x$, $y$, $z$ dimensions. For the voxelization, a fixed grid size and a fixed range are set so that all voxels positions are fixed. For a point in the point cloud, it is assigned to a certain voxel based on its coordinates. Then, in each voxel, we calculate the mean of all points assigned to it and use this mean value as the representative value for that voxel. Finally, a voxelized 3D input is generated and is sent to the 3D Feature Extractor.

\subsection{3D Feature Extractor}
In this part, 3D convolutional layers are used to extract features from voxelized 3D inputs. We use 3D sparse convolutional layer~\cite{yan2018second} and submanifold sparse convolutional layer~\cite{graham2017submanifold} to build our feature extractor.

We propose a lite version of the 3D Feature Extractor. It has fewer residual blocks at the early stage but slightly more channels for each residual block. It only down-samples $z$-axis dimension with a factor of 8. The channel dimension and the $z$-axis dimension are reshaped and combined together after the 3D Feature Extractor. In order to keep the same shape of the output feature map, the number of channels of the feature extraction layers at the final down-sampled scale is decreased accordingly, which also significantly accelerates the 3D Feature Extractor. The detailed structure of the 3D Feature Extractor is shown in Fig.~\ref{fig:backbone}. The resulting feature map is reshaped to form a Bird's Eye View (BEV) pseudo image.

\subsection{Region Proposal Networks}
The Region Proposal Network (RPN) takes a pseudo image as input and employs multiple down-sample and up-sample blocks to output stride 8 feature maps, so the final down-sample factor is 1 for the RPN. The RPN baseline structure is shown in Fig.~\ref{fig:RPN1}. We replace the basic $3 \times 3$ convolutional blocks with self-calibrated convolutions (SC-Conv) \cite{liu2020improving} as shown in Fig.~\ref{fig:RPN2} and~\ref{fig:SCConv}, which helps to enlarge the receptive field for spatial locations and introduces channel-wise and spatial-wise attention with cost-efficiency. The improved RPN structure boosts the detection accuracy with a similar number of parameters and computational costs as the baseline.

\subsection{Anchor-free Head}

\begin{figure*}
\begin{center}

\includegraphics[width=0.95\textwidth]{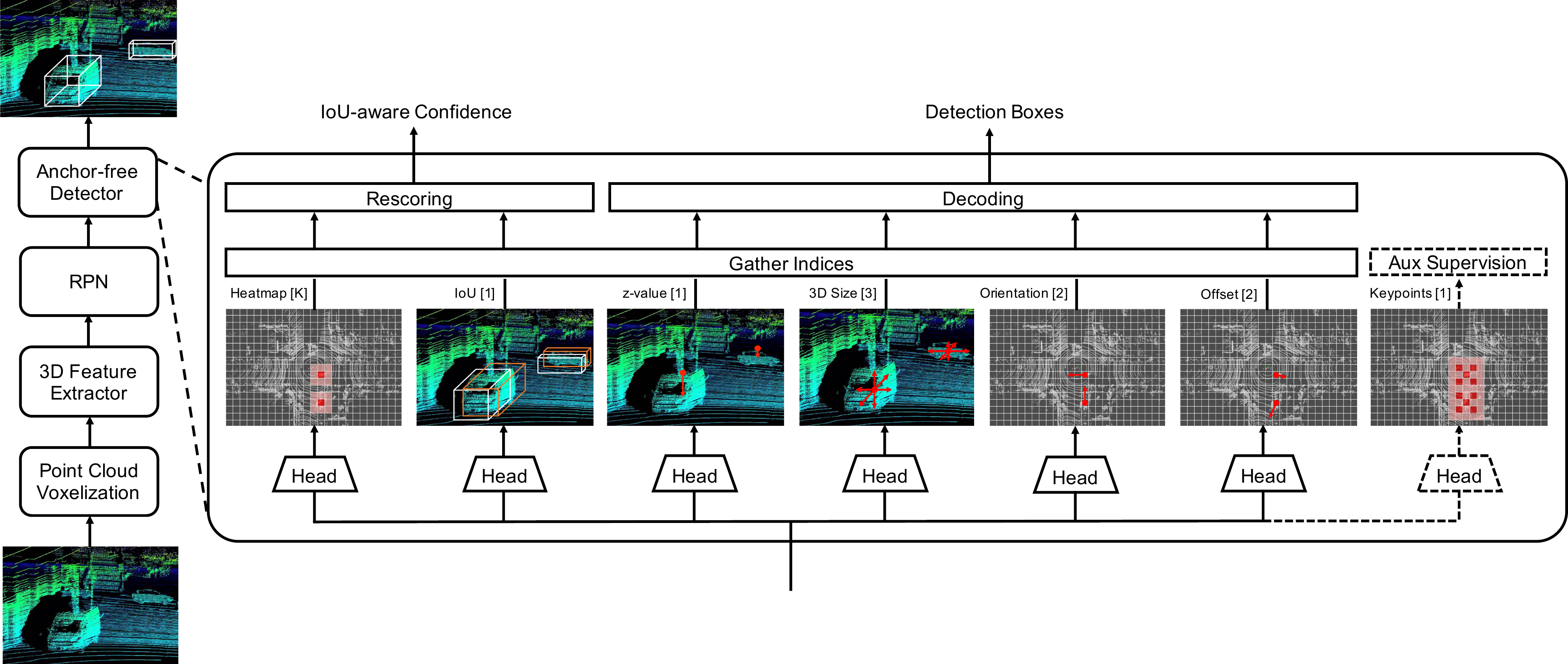}
\end{center}
\caption{The framework of anchor-free one-stage 3D detection (AFDetV2) system. The whole pipeline consists of the Point Cloud Voxelization, 3D Feature Extractor, Region Proposal Networks (RPN) and the Anchor-free Detector. The number in the brackets indicates the output channels in the last convolution layer. $K$ is the number of categories used in the detection. The auxiliary supervision which is turned off at inference is shown in dashed lines. Better viewed in color and zoom in for more details.}
\label{fig:framework}
\end{figure*}
In addition to the five sub-heads in AFDet~\cite{ge2020afdet}, we devise 2 new sub-heads in our AFDetV2 to achieve better accuracy. The 5 sub-heads common to both AFDet and AFDetV2 are the heatmap prediction head, the local offset regression head, the $z$-axis location regression head, the 3D object size regression head and the orientation regression head. The 2 new sub-heads are IoU-aware confidence score prediction and keypoints auxiliary supervision, as shown in Fig.~\ref{fig:framework}. First we will briefly introduce the differences of the shared 5 sub-heads between the AFDet~\cite{ge2020afdet} and AFDetV2. Then we will describe the newly devised sub-heads.

\textbf{Differences of the 5 sub-heads.} For the heatmap head, we enlarge the positive supervision for the heatmap prediction by setting a minimum allowed Gaussian radius~\cite{law2018cornernet} to 2 following~\cite{yin2020center}. For the orientation regression sub-head, we set the regression target to the $sin$ and $cos$ values of the yaw angle of the objects.

\textbf{IoU-aware confidence score prediction.}
In the task of object detection, the classification score is commonly used as the final prediction score. However, it is not a good estimate of the localization accuracy. In this case, boxes with high localization accuracy but low classification scores may be deleted after Non-Maximum Suppression (NMS). Also, the misalignment harms the ranking-based metrics such as Average Precision (AP). To alleviate the misalignment, most of the existing methods adopt an IoU-aware prediction branch as the second stage network ~\cite{jiang2018acquisition,shi2020pv,yin2020center}. However, an additional stage will bring more computational cost and latency to the network. Also, special operator such as RoI Align~\cite{he2017mask, yin2020center} or RoI pooling ~\cite{shi2019pointrcnn,shi2020pv,shi2020points} is required in the second stage network.

In our solution, we adopt an IoU prediction head following~\cite{wu2020iou,zheng2020ciassd}. To incorporate IoU information into confidence score, we recalculate final confidence score by a post-processing function:
\begin{equation}
f = score^{1 - \alpha} * iou ^ {\alpha}
\end{equation}
where $score$ is the original classification score, $iou$ is the predicted IoU and $\alpha$ is a parameter $\in[0, 1]$ that controls the contributions from the classification score and predicted IoU. In our solution, we use 0.68, 0.71, 0.65 for VEHICLE, PEDESTRIAN and CYCLIST respectively. 

After rescoring, the ranking of the predictions takes both the classification confidence and localization accuracy into account. It will lower the confidence of the predictions with higher classification scores but worse localization accuracy, and vice versa.
Our single-stage network runs much faster than most existing two-stage LiDAR detectors while surpassing their detection results, as shown in the leader board in Tab.~\ref{tbl:real-time 3d detection test result}.

\textbf{Keypoints auxiliary supervision.} Inspired by the corner classification in~\cite{wang2020centernet3d}, we add a sub-head of keypoint prediction as an auxiliary supervision in the detection. Specifically, another heatmap that predicts 4 corners and the center of every object in BEV is added during training. We use the same method to draw the objects on the heatmap as the heatmap prediction sub-head but with a halved radius and minimum Gaussian radius~\cite{law2018cornernet} set to 1. It should be noted that this sub-head does not affect the inference speed as the sub-head is disabled at inference.

\subsection{Loss}
Following~\cite{ge2020afdet, ding20201st, wang20201st}, we use the focal loss~\cite{lin2017focal, law2018cornernet, yin2020center} for the heatmap prediction and keypoints auxiliary supervision. $L_1$ loss is employed in the local offset head, the $z$-axis location head, the 3D object size head, and orientation regression. For IoU prediction branch, we encode the target IoU as $(2 * iou - 0.5)$ $\in$ $[-1, 1]$ following~\cite{zheng2020ciassd}, where $iou$ is the 3D axis-aligned IoU between the predicted box and associated ground truth box. Then smooth $L_1$ loss is employed to regress the encoded IoU.

The overall training objective is
\begin{equation}
\begin{split}
    \mathcal{L} = \mathcal{L}_{heat} + \lambda_{off}\mathcal{L}_{off} + \lambda_{z}\mathcal{L}_{z} + \lambda_{size}\mathcal{L}_{size} +\\ \lambda_{ori}\mathcal{L}_{ori} + \lambda_{iou}\mathcal{L}_{iou} + \lambda_{kps}\mathcal{L}_{kps}
\end{split}
\label{eq:loss}
\end{equation}

where $\lambda$ represents the weight for each sub-task. For all regression sub-heads including local offset, $z$-axis location, 3D size, orientation, and IoU prediction, we only regress $N$ foreground objects that are in the detection range.

\section{Experiments}
\subsection{Point Clouds Densification and Data Inputs}
Accumulating LiDAR sweeps is a simple but effective method to utilize temporal information and densify LiDAR point cloud~\cite{caesar2019nuscenes, ding20201st}. To distinguish points from different sweeps, the time lag $\mathit{\Delta t}$ is attached to the point cloud as an additional attribute. In our solution, we use the past one frame combined with the current frame as our input point cloud.

Waymo Open Dataset~\cite{sun2020scalability} provides five types of LiDAR sweeps. To reduce the latency, we only utilize the first and second returns of the top LiDAR. Specifically, considering frame densification, our input format should be $\mathit{(x,y,z, reflectance, elongation, \Delta t)}$, while $\mathit{\Delta t}$ being the time lag as mentioned above.

\subsection{Data Augmentation}
We use data augmentation strategy following~\cite{yan2018second,zhu2019class,ding20201st}. First, we generate an annotation database containing labels and the associated point cloud data. During training, we randomly select 15, 10 and 10 ground truth samples for vehicle, pedestrian and cyclist respectively, and place them into the current frame. Second, we do random flipping along $x,y$-axis, global rotation with a uniform distribution of $\mathcal{U}\left ( -\frac{\pi}{4}, \frac{\pi}{4} \right )$, global scaling of $\mathcal{U}\left ( 0.95, 1.05 \right )$ and global translation along $x,y,z$-axis of $\mathcal{U}\left ( -0.2m, 0.2m \right )$. Third, random rotation noise with a uniform distribution of  $\mathcal{U}\left ( -\frac{\pi}{20}, \frac{\pi}{20} \right )$ and random  location noise with a Gaussian distribution $\mathcal{N}\left ( 0.0, ~0.1 \right )$ are applied to each instance.

\subsection{Stochastic Weights Averaging}
Stochastic Weights Averaging (SWA) is developed in ~\cite{izmailov2018averaging} for improving generalization in deep networks. Inspired by VarifocalNet~\cite{zhang2020varifocalnet, zhang2020swa}, we equipped SWA to our model and achieved $\sim$0.2 mAPH/L2 improvement as shown in Tab.~\ref{tbl:swa results}. Specifically, we trained 5 extra epochs with 1/10 of the original learning rate. We adopted a cyclical learning rate scheduler with one-cycle policy~\cite{one_cycle}. In each epoch, after a warming up from $3 \times 10^{-5}$ to  $3 \times 10^{-4}$, the learning rate decreases at each iteration from  $3 \times 10^{-4}$ to $3 \times 10^{-9}$. Finally, we calculated the average of 5 checkpoints and trained an extra epoch to fix the Batch Normalization~\cite{pmlr-v37-ioffe15} parameters.
\begin{table}[t]
\begin{center}
\setlength\tabcolsep{3pt}
\begin{tabular}{l|c|c|c|c}
\hline
  Checkpoints     &  Vehicle   &Pedestrian  &Cyclist  &All\\
\hline\hline
 Baseline &\multicolumn{1}{|c|}{70.26} &\multicolumn{1}{|c|}{71.17} &\multicolumn{1}{|c|}{72.20} &\multicolumn{1}{|c}{71.21}\\
\hline
 Epoch 1 &\multicolumn{1}{|c|}{70.38} &\multicolumn{1}{|c|}{71.70} &\multicolumn{1}{|c|}{71.72} &\multicolumn{1}{|c}{71.27}\\	
  Epoch 2 &\multicolumn{1}{|c|}{70.47} &\multicolumn{1}{|c|}{71.22} &\multicolumn{1}{|c|}{72.23} &\multicolumn{1}{|c}{71.31}\\
   Epoch 3 &\multicolumn{1}{|c|}{70.45} &\multicolumn{1}{|c|}{71.70} &\multicolumn{1}{|c|}{71.75} &\multicolumn{1}{|c}{71.30}\\
    Epoch 4 &\multicolumn{1}{|c|}{70.41} &\multicolumn{1}{|c|}{71.74} &\multicolumn{1}{|c|}{71.82} &\multicolumn{1}{|c}{71.32}\\
     Epoch 5 &\multicolumn{1}{|c|}{70.46} &\multicolumn{1}{|c|}{71.75} &\multicolumn{1}{|c|}{71.92} &\multicolumn{1}{|c}{71.38}\\

\hline
 After SWA &\multicolumn{1}{|c|}{70.54} &\multicolumn{1}{|c|}{71.72} &\multicolumn{1}{|c|}{72.00} &\multicolumn{1}{|c}{71.42}\\

\hline
\end{tabular}
\end{center}
\caption{Improvement of model's performance after SWA. Performance of each SWA checkpoints are also listed. Values in this table are results of the mAPH/L2 evaluation metric. Models are tested on 1/5 validation set.} 
\label{tbl:swa results}
\end{table}

\subsection{Improvements on Latency}

To reduce the inference latency, we exploited GPU on the voxelization and the conversion of range data to point features.

\textbf{Voxelization with GPU.}
Voxelization is composed of two processes: first, each point is quantized with the grid cell size $(g_x, g_y, g_z)$ to the corresponding voxel $(c_x, c_y, c_z)$, where $c_x=\lfloor x/g_x\rfloor, c_y= \lfloor y/g_y\rfloor, c_z=\lfloor z/g_z\rfloor$ are the voxel coordinates, and $\lfloor\cdot\rfloor$ denotes the $floor()$ operation; and second, an average of all the points in a voxel is calculated to form the feature vector for each voxel, \ie $F_i(d) = \frac{1}{N_i}\sum\limits_{k\in V_i} P_k(d), d \in D$, where $N_i$ is the number of points in the corresponding voxel, $P_k$ is a point, $V_i$ is the corresponding voxel, and $D$ is the dimension of the feature vector.

To accelerate voxelization, we implemented both processes in GPU, as follows:

(1)	As an initialization step, we allocate two buffers in the GPU memory, one for the feature values in dense voxels in a grid of $(v_x,  v_y,  v_z)$, denoted as $F_g$, and the other for the count of points in a voxel, also in the same grid,  denoted as $N_g$.

(2)	For each point $P_i$, we calculate the corresponding voxel coordinate $(c_x, c_y, c_z)$ and do the atomic accumulation in $F_g(c_x, c_y, c_z)$ and $N_g(c_x, c_y, c_z)$. At the same time, we store the voxel index $i_v$ for $P_i$ to a list $I_p$, where $i_v = c_x + c_y \cdot v_x + c_z \cdot v_x \cdot y_y$. It should be noted that in this implementation, we do not limit the number of points per voxel.

(3)	Transfer the voxel index list $I_p$ from GPU memory to host memory. Note that the size of $I_p$ is the number of points in the given point cloud.

(4)	Using CPU, we remove the duplicates of voxel index in $I_p$ and obtain a list of unique voxel index, denoted as $I_u$.

(5)	Transfer $I_u$ from CPU memory to GPU. Note that $[I_u] \leqslant [I_p]$, where $[\cdot]$ denotes the size of a set.

(6)	Gather the cumulative features and the counts in $F_g$ and $N_g$ respectively on GPU, and calculate $F_i$ for each individual voxel indexed in $I_u$.

\textbf{Range data conversion with GPU.} 
A range data $R_i$ is a vector containing $(range, intensity, elongation, x, y, z)$. We convert it to a point $P_i$ containing $(x, y, z, intensity_{clamp}, elongation)$ before voxelization, where $intensity_{clamp}$ is $intensity$ clamped at $1.5$. For the case of densification with previous frames, $P_i$ contains $(x, y, z, intensity_{clamp}, elongation, \Delta t)$, where $\Delta t$ is the time interval between the current frame and the previous frame. 

To convert an $R_i$ to $P_i$, we utilize both CPU and GPU in the following steps:

(1) Exclude the invalid $R_i$ by checking the $range$ value. This is done on CPU.

(2) Transfer the valid range data to GPU memory. For densification with previous frames, also transfer the transform matrices (\ie the extrinsics).

(3) On GPU, for each $R_i$, permute the entries as described above, and apply the extrinsics to the $(x, y, z)$ values of the previous frames for densification.

\subsection{Experiment Settings}
\label{exp_settings}
Only the point clouds from the top LiDAR were used during training and inference for the purpose of acceleration. The maximum number of objects was set to $500$. For all of our models, we set the max point per voxel to $15$, max voxel num to $300,000$ during training. We did not set a limit for the max number of points at inference. We used AdamW~\cite{loshchilov2018decoupled} optimizer with one-cycle policy~\cite{one_cycle}. We set the learning rate max to $3\times10^{-3}$, division factor to 10, momentum ranges from 0.95 to 0.85, fixed weight decay to 0.01 to achieve convergence. we set all $\lambda$ in Eqn.~\ref{eq:loss} to 2.0.

To quickly verify our idea, we sampled the validation set every 5 frames according to their timestamps. Unless explicitly indicated, all models used for the ablation studies, including Tab.~\ref{tbl:swa results} and Tab.~\ref{tbl:ablation study}, were tested on 1/5 validation data. 

 AFDetV2-Base and AFDetV2-Lite were trained and tested within range [(-75.2, 75.2), (-75.2, 75.2),  (-2, 4)] in respect of $x,y,z$-axis as shown in Tab.~\ref{tbl:model training}. AFDetV2 was trained within range [(-75.2, 75.2), (-73.6, 73.6), (-2, 4)], while enlarging the detection range to [(-80, 80), (-76.16, 76.16), (-2, 4)] at inference. For AFDetV2-Base, we first trained 10 epochs on training data and finetuned 36 epochs on the whole trainval data with [0.1, 0.1, 0.15] grid size with respect to the $x,y,z$-axis. For AFDetV2, we adjusted the grid size to [0.1, 0.08, 0.15] and finetuned another 36 epochs on the whole trainval set based on AFDetV2-Base. Two frames of point clouds were utilized for densification in AFDetV2-Base and AFDetV2, while AFDetV2-Lite only exploited a single frame of the point cloud. AFDetV2-Lite was only finetuned 18 epochs on trainval set after being trained on training data for 10 epochs with grid size [0.1, 0.1, 0.15]. AFDetV2-Base and AFDetV2-Lite were trained with 10 samples per GPU while AFDetV2 was trained with 8 samples per GPU. For the offset regression head, we set the regression radius to 0. Besides, we replaced max pooling with class-specific NMS for better average precision. In our solution, we set IoU threshold to 0.8, 0.55, 0.55 for VEHICLE, PEDESTRIAN and CYCLIST respectively. Models were trained with Nvidia V100 GPUs.

To reduce the latency, the following improvements are adopted at inference: 

(1)	Range images were converted to point clouds in Cartesian coordinate on GPU. 

(2)	Voxelization process was executed with GPU. 

(3)	All layers were cast to half-precision (FP16) except for the last layers in each sub-heads.

(4)	Batch Normalization~\cite{pmlr-v37-ioffe15} parameters were merged into 3D Sparse Convolution and SubManifold 3D Sparse Convolution layers in 3D Feature Extractor.

(5)	Keypoints auxiliary branch was disabled.


\begin{table}[h]
\begin{center}

\vspace{5pt}
\setlength\tabcolsep{2pt}
\begin{tabular}{l|c|c c|c c|c c}
\hline
\multirow{2}{*}{Models} &Num. of  &\multicolumn{2}{c|}{Training} &\multicolumn{2}{c|}{Inference}  &\multicolumn{2}{c}{Grid} \\
&Frames &\multicolumn{2}{c|}{Range} &\multicolumn{2}{c|}{Range} &\multicolumn{2}{c}{Size}\\
\hline
\hline
AFDetV2       &2  &75.2 &73.6 &80.0 &76.16 &0.1 &0.08 \\
AFDetV2-Base  &2  &75.2 &75.2 &75.2 &75.2  &0.1 &0.1  \\
AFDetV2-Lite  &1  &75.2 &75.2 &75.2 &75.2  &0.1 &0.1  \\
\hline
\end{tabular}
\end{center}
\caption{The configurations of our models. Values in ``training range'',``inference range'' and ``grid size'' columns are in meters. The first value in a column refers to  $x$-axis, while the second value refers to $y$-axis. Please refer to Sec.~\ref{exp_settings} for more details.}
\label{tbl:model training}
\end{table}
\begin{figure*}
\begin{center}

\includegraphics[width=1\textwidth]{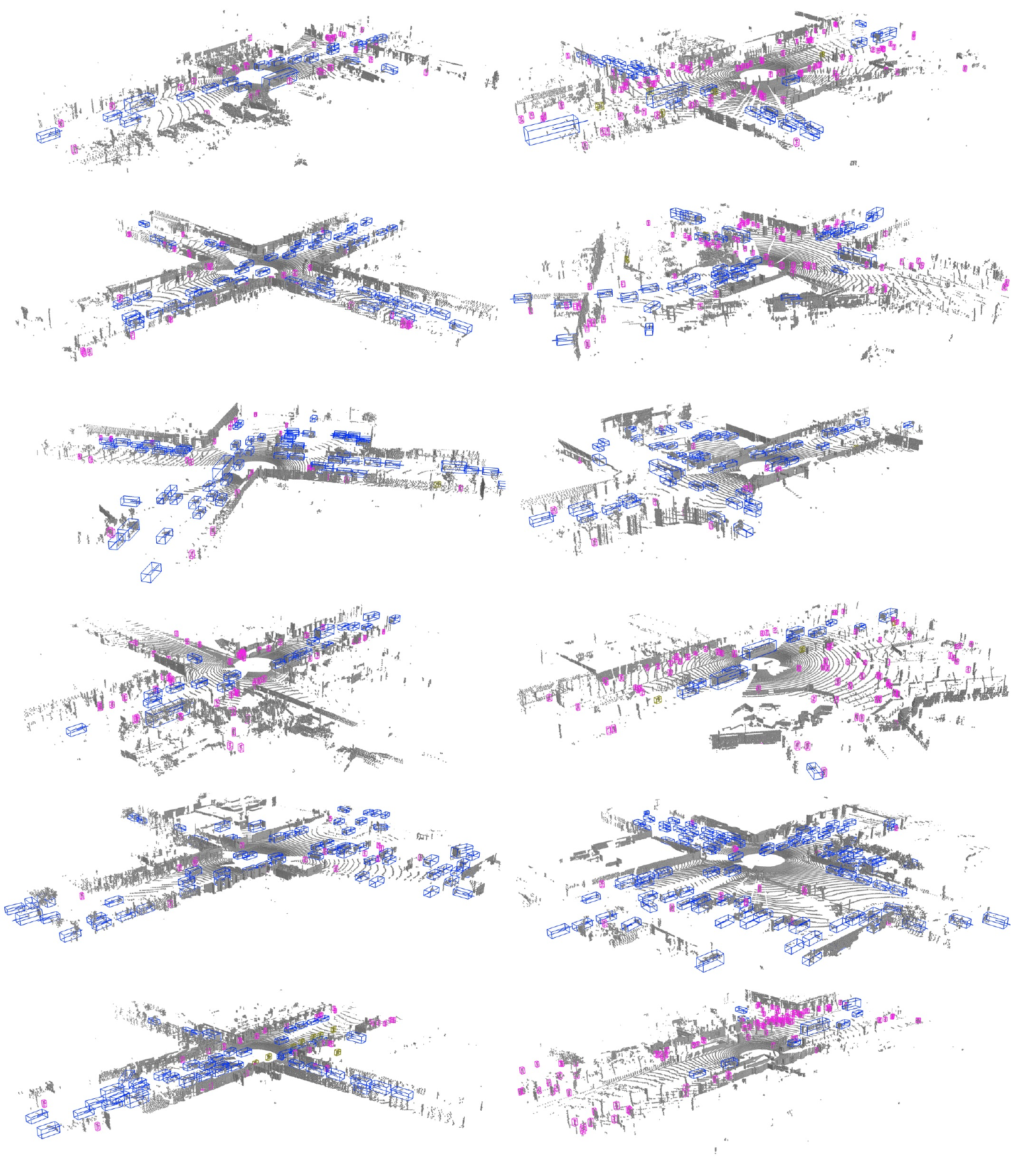}
\end{center}
\caption{The examples of results on testing set of Real-time 3D detection track, only bounding boxes with score larger than 0.50 are visualized. Additional NMS is conducted for better visualization.}
\label{fig:Final result 1}
\end{figure*}

\section{Results}
\subsection{Performance on the Test Set}
The final results on the official real-time 3D detection leaderboard are shown in Tab.~\ref{tbl:real-time 3d detection test result}. As seen from Tab.~\ref{tbl:real-time 3d detection test result}, our AFDetV2 submission achieved the \nth{1} place of the real-time 3D detection challenge and reached 73.12 mAPH/L2 with the latency of 60.06 ms. Models that are eligible for the title of ``Most Efficient Model''  are listed in Tab.~\ref{tbl:most efficient result}, where the primary evaluation metric is the models' latency. Our AFDetV2-Base submission was given the title of ``Most Efficient Model'' by the challenge sponsor and reached 72.57 mAPH/L2 with the latency of 55.86 ms. Our AFDetV2-Lite submission achieved 69.95 mAPH/L2, only 0.05 below 70.00, but with the latency of 46.90 ms, which is 8.94 ms faster than our ``Most Efficient Model''.
\begin{table*}[h]
\begin{center}

\vspace{5pt}
\setlength\tabcolsep{2pt}
\begin{tabular}{l|c|c|c|cc|c}
\hline
 \multirow{2}{*}{Model}  &\multicolumn{1}{|c|}{Keypoints Aux.} &\multicolumn{1}{|c|}{Refined Data} &\multicolumn{1}{|c|}{SC-Conv} &\multicolumn{1}{|c}{ALL}\\
 &Supervision &Augmentation &RPN &mAPH/L2 \\
\hline
\hline
\multirow{4}{*}{w/o IoU branch}  &\xmark &\xmark &\xmark &65.31 \\
 &\cmark &\xmark &\xmark &65.55 \\
 &\cmark &\cmark &\xmark &65.65 \\
 &\cmark &\cmark &\cmark &66.71 \\
\hline
\multirow{2}{*}{w/ IoU branch}  &\cmark &\xmark &\xmark &68.13 \\
 &\cmark &\cmark &\cmark &69.50 \\

\hline
\end{tabular}
\end{center}
\caption{The improvements of accuracy brought by different modules. All models were trained with densified point clouds. Results are shown in terms of mAPH/L2 for all classes, \ie VEHICLE, PEDESTRIAN, CYCLIST. All models were tested on 1/5 validation set.} 
\label{tbl:ablation study}
\end{table*}
\begin{table*}[h]
\begin{center}
\resizebox{1\textwidth}{!}{%
\vspace{5pt}
\setlength\tabcolsep{2pt}
\begin{tabular}{l|c c|c c|c c|c c|c c}
\hline
 \multirow{2}{*}{Models}  &\multicolumn{2}{|c|}{VEHICLE} &\multicolumn{2}{|c}{PEDESTRIAN} &\multicolumn{2}{|c}{CYCLIST} &\multicolumn{2}{|c}{ALL}
 &\multicolumn{1}{|c}{Latency}\\
  &mAP/L2 &mAPH/L2 &mAP/L2 &mAPH/L2 &mAP/L2 &mAPH/L2 &mAP/L2 &mAPH/L2 &ms\\
\hline
\hline
AFDetV2 (Ours) &74.30  &73.89 &\textbf{75.47} &\textbf{72.41} &\textbf{74.05}  &\textbf{73.04} &\textbf{74.60} &\textbf{73.12} &60.06  \\
CenterPoint++  &\textbf{75.47}  &\textbf{75.05} &75.13 &72.41 &72.04  &71.01 &74.22 &72.82 &57.12  \\
AFDetV2-Base (Ours) &73.89  &73.46 &75.34 &72.29 &72.96  &71.97 &74.06 &72.57 &55.86   \\
X\_Autonomous3D  &74.04  &73.60 &72.29 &68.27 &70.55  &69.50 &72.29 &70.46 &68.42   \\
AFDetV2-Lite (Ours) &72.98  &72.55 &73.71 &68.61 &69.84  &68.67 &72.18 &69.95  &\textbf{46.90} \\
\hline
\end{tabular}
}
\end{center}
\caption{Top five submissions of the Real-time 3D Detection. Only models that run faster than 70 ms are qualified for awards. mAPH/L2 of all classes, \ie VEHICLE, PEDESTRIAN, CYCLIST, is the primary metric for the award evaluation.} 
\label{tbl:real-time 3d detection test result}
\end{table*}
\begin{table}[t]
\begin{center}
\setlength\tabcolsep{3pt}
\begin{tabular}{l|c c|c}
\hline
 \multirow{2}{*}{Models}  &\multicolumn{2}{|c}{ALL} &\multicolumn{1}{|c}{Latency}\\
  &mAP/L2 &mAPH/L2 &ms\\
\hline\hline
AFDetV2-Base (Ours) &74.06 &72.57 &\textbf{55.86}   \\
CenterPoint++  &74.22 &72.82 &57.12  \\
AFDetV2 (Ours) &\textbf{74.60} &\textbf{73.12} &60.06  \\
X\_Autonomous3D  &72.29 &70.46 &68.42   \\

\hline
\end{tabular}
\end{center}
\caption{Top four submissions of the Real-time 3D Detection ``Most Efficient Model'' title. Only models with mAPH/L2 greater than 70 are qualified for this title. Latency is the primary metric for the ``Most Efficient Model'' title evaluation.} 
\label{tbl:most efficient result}
\end{table}

\subsection{Ablation Studies}
To study the effect of each module deployed in our solution, we conducted ablation experiments on the 1/5 validation set as shown in Tab.~\ref{tbl:ablation study}. We divided the ablation study into two sections: one for the case without the IoU branch, and the other with the IoU branch. From the top part of Tab.~\ref{tbl:ablation study}, we can see that without the IoU branch, the SC-Conv~\cite{liu2020improving} led to an increment of detection performance of 1.06 mAPH/L2. The keypoints auxiliary loss contributed about 0.24 mAPH/L2. From the bottom part of Tab.~\ref{tbl:ablation study}, we observe that the IoU branch can significantly increase the accuracy. With the IoU branch, the accuracy increased by 2.79 mAPH/L2 compared to the case of without the IoU branch, when all the other modules were deployed.

\section{Conclusion}
We have shown a real-time, single-stage and anchor-free 3D object detection model named AFDetV2. This model is composed of a lite 3D Feature Extractor, a refined SC-Conv RPN and an anchor-free head with a novel IoU-aware branch and a keypoints auxiliary branch. To reduce the inference latency, we also devised a GPU-based voxelization method. Thanks to these innovations, AFDetV2 achieved the \nth{1} Place award for Real-time 3D Detection and the title of ``Most Efficient Model'' of the Waymo Open Dataset Challenges 2021.

{\small
\bibliographystyle{ieee_fullname}
\bibliography{egbib}

\begin{thebibliography}{10}\itemsep=-1pt

\bibitem{caesar2019nuscenes}
Holger Caesar, Varun Bankiti, Alex~H Lang, Sourabh Vora, Venice~Erin Liong,
  Qiang Xu, Anush Krishnan, Yu Pan, Giancarlo Baldan, and Oscar Beijbom.
\newblock nuscenes: A multimodal dataset for autonomous driving.
\newblock In {\em CVPR}, 2020.

\bibitem{ding20201st}
Zhuangzhuang Ding, Yihan Hu, Runzhou Ge, Li Huang, Sijia Chen, Yu Wang, and Jie
  Liao.
\newblock 1st place solution for waymo open dataset challenge--3d detection and
  domain adaptation.
\newblock {\em arXiv preprint arXiv:2006.15505}, 2020.

\bibitem{ettinger2021large}
Scott Ettinger, Shuyang Cheng, Benjamin Caine, Chenxi Liu, Hang Zhao, Sabeek
  Pradhan, Yuning Chai, Ben Sapp, Charles Qi, Yin Zhou, et~al.
\newblock Large scale interactive motion forecasting for autonomous driving:
  The waymo open motion dataset.
\newblock {\em arXiv preprint arXiv:2104.10133}, 2021.

\bibitem{ge2020afdet}
Runzhou Ge, Zhuangzhuang Ding, Yihan Hu, Yu Wang, Sijia Chen, Li Huang, and
  Yuan Li.
\newblock Afdet: Anchor free one stage 3d object detection.
\newblock In {\em CVPR Workshops}, 2020.

\bibitem{graham2017submanifold}
Benjamin Graham and Laurens van~der Maaten.
\newblock Submanifold sparse convolutional networks.
\newblock {\em arXiv preprint arXiv:1706.01307}, 2017.

\bibitem{one_cycle}
Sylvain Gugger.
\newblock The 1cycle policy.
\newblock \url{https://sgugger.github.io/the-1cycle-policy.html}, 2018.

\bibitem{he2017mask}
Kaiming He, Georgia Gkioxari, Piotr Doll{\'a}r, and Ross Girshick.
\newblock Mask r-cnn.
\newblock In {\em ICCV}, 2017.

\bibitem{pmlr-v37-ioffe15}
Sergey Ioffe and Christian Szegedy.
\newblock Batch normalization: Accelerating deep network training by reducing
  internal covariate shift.
\newblock In {\em ICML}, 2015.

\bibitem{izmailov2018averaging}
Pavel Izmailov, Dmitrii Podoprikhin, Timur Garipov, Dmitry Vetrov, and
  Andrew~Gordon Wilson.
\newblock Averaging weights leads to wider optima and better generalization.
\newblock In {\em UAI}, 2018.

\bibitem{jiang2018acquisition}
Borui Jiang, Ruixuan Luo, Jiayuan Mao, Tete Xiao, and Yuning Jiang.
\newblock Acquisition of localization confidence for accurate object detection.
\newblock In {\em ECCV}, 2018.

\bibitem{law2018cornernet}
Hei Law and Jia Deng.
\newblock Cornernet: Detecting objects as paired keypoints.
\newblock In {\em ECCV}, 2018.

\bibitem{lin2017focal}
Tsung-Yi Lin, Priya Goyal, Ross Girshick, Kaiming He, and Piotr Doll{\'a}r.
\newblock Focal loss for dense object detection.
\newblock In {\em CVPR}, 2017.

\bibitem{liu2020improving}
Jiang-Jiang Liu, Qibin Hou, Ming-Ming Cheng, Changhu Wang, and Jiashi Feng.
\newblock Improving convolutional networks with self-calibrated convolutions.
\newblock In {\em CVPR}, 2020.

\bibitem{loshchilov2018decoupled}
Ilya Loshchilov and Frank Hutter.
\newblock Decoupled weight decay regularization.
\newblock In {\em ICLR}, 2019.

\bibitem{shi2020pv}
Shaoshuai Shi, Chaoxu Guo, Li Jiang, Zhe Wang, Jianping Shi, Xiaogang Wang, and
  Hongsheng Li.
\newblock Pv-rcnn: Point-voxel feature set abstraction for 3d object detection.
\newblock In {\em CVPR}, 2020.

\bibitem{shi2019pointrcnn}
Shaoshuai Shi, Xiaogang Wang, and Hongsheng Li.
\newblock Pointrcnn: 3d object proposal generation and detection from point
  cloud.
\newblock In {\em CVPR}, 2019.

\bibitem{shi2020points}
Shaoshuai Shi, Zhe Wang, Jianping Shi, Xiaogang Wang, and Hongsheng Li.
\newblock From points to parts: 3d object detection from point cloud with
  part-aware and part-aggregation network.
\newblock {\em TPAMI}, 2020.

\bibitem{sun2020scalability}
Pei Sun, Henrik Kretzschmar, Xerxes Dotiwalla, Aurelien Chouard, Vijaysai
  Patnaik, Paul Tsui, James Guo, Yin Zhou, Yuning Chai, Benjamin Caine, et~al.
\newblock Scalability in perception for autonomous driving: Waymo open dataset.
\newblock In {\em CVPR}, 2020.

\bibitem{wang2020centernet3d}
Guojun Wang, Bin Tian, Yunfeng Ai, Tong Xu, Long Chen, and Dongpu Cao.
\newblock Centernet3d: An anchor free object detector for autonomous driving.
\newblock {\em arXiv preprint arXiv:2007.07214}, 2020.

\bibitem{wang20201st}
Yu Wang, Sijia Chen, Li Huang, Runzhou Ge, Yihan Hu, Zhuangzhuang Ding, and Jie
  Liao.
\newblock 1st place solutions for waymo open dataset challenges--2d and 3d
  tracking.
\newblock {\em arXiv preprint arXiv:2006.15506}, 2020.

\bibitem{wu2020iou}
Shengkai Wu, Xiaoping Li, and Xinggang Wang.
\newblock Iou-aware single-stage object detector for accurate localization.
\newblock {\em Image and Vision Computing}, 97:103911, 2020.

\bibitem{yan2018second}
Yan Yan, Yuxing Mao, and Bo Li.
\newblock Second: Sparsely embedded convolutional detection.
\newblock {\em Sensors}, 18(10):3337, 2018.

\bibitem{yin2020center}
Tianwei Yin, Xingyi Zhou, and Philipp Kr{\"a}henb{\"u}hl.
\newblock Center-based 3d object detection and tracking.
\newblock In {\em CVPR}, 2021.

\bibitem{zhang2020swa}
Haoyang Zhang, Ying Wang, Feras Dayoub, and Niko S{\"u}nderhauf.
\newblock Swa object detection.
\newblock {\em arXiv preprint arXiv:2012.12645}, 2020.

\bibitem{zhang2020varifocalnet}
Haoyang Zhang, Ying Wang, Feras Dayoub, and Niko S{\"u}nderhauf.
\newblock Varifocalnet: An iou-aware dense object detector.
\newblock In {\em CVPR}, 2021.

\bibitem{zheng2020ciassd}
Wu Zheng, Weiliang Tang, Sijin Chen, Li Jiang, and Chi-Wing Fu.
\newblock Cia-ssd: Confident iou-aware single-stage object detector from point
  cloud.
\newblock In {\em AAAI}, 2021.

\bibitem{zhou2018voxelnet}
Yin Zhou and Oncel Tuzel.
\newblock Voxelnet: End-to-end learning for point cloud based 3d object
  detection.
\newblock In {\em CVPR}, 2018.

\bibitem{zhu2019class}
Benjin Zhu, Zhengkai Jiang, Xiangxin Zhou, Zeming Li, and Gang Yu.
\newblock Class-balanced grouping and sampling for point cloud 3d object
  detection.
\newblock {\em arXiv preprint arXiv:1908.09492}, 2019.

\end{thebibliography}
}

\end{document}